\documentclass[twocolumn,letterpaper]{IEEEAerospaceCLS}  % only supports two-column, letterpaper format
\usepackage[font=small]{caption}
\usepackage[]{amsmath,amssymb,dsfont,color,graphicx,subcaption,mathtools,nccmath,bm,enumitem,tabularx}
\usepackage{url}
\usepackage[space]{cite}

\newcolumntype{b}{X}
\newcolumntype{s}{>{\hsize=.45\hsize}X}
\newcolumntype{t}{>{\hsize=.25\hsize}X}
\usepackage{makecell} \setcellgapes{2pt}

\begin{document}
\title{Designing ReachBot: System Design Process with a Case Study of a Martian Lava Tube Mission}

\author{%
Stephanie Newdick\\ 
Dept. of Aero. \& Astro.\\
Stanford University\\
496 Lomita Mall\\
Stanford, CA 94305\\
snewdick@stanford.edu
\and
Tony G. Chen\\ 
Dept. of Mech. Engineering\\
Stanford University\\
440 Escondido Mall\\
Stanford, CA 94305\\
agchen@stanford.edu
\and
Benjamin Hockman\\
Robotic Mobility Group\\
Jet Propulsion Laboratory\\
California Institute of Technology\\
Pasadena, CA 91109\\
benjamin.j.hockman@jpl.nasa.gov\\
\and
Edward Schmerling\\
Dept. of Aero. \& Astro.\\
Stanford University\\
496 Lomita Mall\\
Stanford, CA 94305\\
schmrlng@stanford.edu
\and
Mark R. Cutkosky\\ 
Dept. of Mech. Engineering\\
Stanford University\\
440 Escondido Mall\\
Stanford, CA 94305\\
cutkosky@stanford.edu
\and 
Marco Pavone\\ 
Dept. of Aero. \& Astro.\\
Stanford University\\
496 Lomita Mall\\
Stanford, CA 94305\\
pavone@stanford.edu
%%%% IMPORTANT: Use the correct copyright information--IEEE, Crown, or U.S. government. %%%%%
\thanks{\footnotesize 978-1-6654-9032-0/23/$\$31.00$ \copyright2023 IEEE}           
}

\maketitle

\thispagestyle{plain}
\pagestyle{plain}

\maketitle

\thispagestyle{plain}
\pagestyle{plain}

%%%%%%%%%%%%%%%%%%%%%%%%%%%%%%%%%%%%%%%%%%%%%%%%%%%%%%%%%%%%%%%%%%%%%%%%%%%%%%%%
\begin{abstract}
In this paper we present a trade study-based method to optimize the architecture of ReachBot, a new robotic concept that uses deployable booms as prismatic joints for mobility in environments with adverse gravity conditions and challenging terrain.
Specifically, we introduce a design process wherein we analyze the compatibility of ReachBot's design with its mission. 
We incorporate terrain parameters and mission requirements to produce a final design optimized for mission-specific objectives.
ReachBot's design parameters include (1) number of booms, (2) positions and orientations of the booms on \mbox{ReachBot's} chassis, (3) boom maximum extension, (4) boom cross-sectional geometry, and (5) number of active/passive degrees-of-freedom at each joint. Using first-order approximations, we analyze the relationships between these parameters and various performance metrics including stability, manipulability, and mechanical interference. 
We apply our method to a mission where ReachBot navigates and gathers data from a martian lava tube. The resulting design is shown in Fig.~\ref{fig:final_config}.
\end{abstract}
\tableofcontents    

\begin{figure}
    \centering
    %MRC Very cool image. Now it looks even more like
    %a harvestman arachind :-)
    % https://en.wikipedia.org/wiki/Opiliones
    \includegraphics[width=\columnwidth]{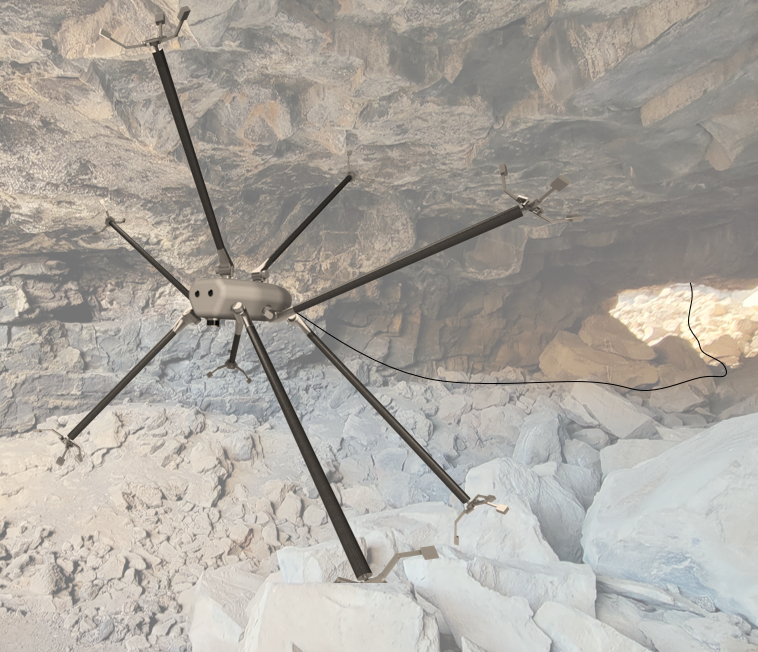}
    \caption{ReachBot exploring a martian lava tube. The final configuration for this mission profile has eight booms positioned at the sides and top of the body with a leading LiDAR, trailing tether, and science suite aimed at the surface below.}
    \label{fig:final_config}
\end{figure}
%%%%%%%%%%%%%%%%%%%%%%%%%%%%%%%%%%%%%%%%%%%%%%%%%%%%%%%%%%%%%%%%%%%%%%%%%%%%%%%%
\section{Introduction} \label{sec:intro}
%%%%%%%%%%%%%%%%%%%%%%%%%%%%%%%%%%%%%%%%%%%%%%%%%%%%%%%%%%%%%%%%%%%%%%%%%%%%%%%%
The scientific interest in exploring novel environments requires robots capable of versatile mobility in challenging and unpredictable terrains.
In particular, NASA's interest in understanding the history of the solar system requires exploring caves, cliffs, and other rocky terrain on planetary bodies like Mars and the moon~\cite{NRC2011}.
Additionally, the search for habitable conditions and astrobiological signatures points to icy worlds such as Enceladus and Europa as targets for further investigation~\cite{mckay2014follow}.
These scientifically driven missions introduce new challenges for robotic mobility.
For example, navigating caves and cliffs calls for a robot with anchored mobility under adverse gravity, that is, where the robot must adhere to steep or overhanging surfaces~\cite{LapotreORourkeEtAl2020}, but loose material, obstacles, and crevasses in these environments make the anchor points sparse and the terrain unpredictable. 
Additionally, features on the surface of icy moons are largely unknown, reiterating the importance of mobility in unpredictable terrain.

ReachBot is a new robot concept for versatile mobility in challenging environments. 
Adverse gravity conditions and unpredictable terrains expose a technological need in traditional robots: operations often require careful anchor or footstep placement, but the limited workspace and manipulation capabilities of existing robot designs impede mobility in these environments.
As debuted in our past works~\cite{SchneiderBylardEtAl2022,ChenMillerEtAl2022,NewdickOngoleEtAl2023}, ReachBot uses multiple
deployable booms as prismatic joints to achieve a large workspace which it leverages to navigate challenging terrain. Through their lightweight, compactable structure, these booms also reduce mass and complexity compared to traditional rigid-link articulated-arm designs.
ReachBot's large workspace increases the number of available anchor points and accessible footsteps, a notable advantage when navigating environments with sparse anchor points, obstructions, or large variations in terrain.

The combination of climbing or walking with long reach makes the ReachBot paradigm amenable to a variety of different environments, but the relative importance of design features depends on the specific mission. For example, in missions with surrounding walls that enclose the robot, 
ReachBot can act like a cable-driven robot by keeping its booms in tension, requiring booms with strong tensile strength but negligible compression requirements. 
Conversely, in a mission to navigate a gently curved surface, ReachBot acts as a legged walker and demands that its booms are loaded in compression.
These different mission scenarios both take advantage of ReachBot's long booms, but load them somewhat differently, leading to different optimal configurations. More generally, for any mission,
we must balance operational priority, such as maximizing scientific value, with the technical feasibility of a novel mobile manipulation platform.

\textit{Statement of Contributions:}
This paper presents a method for designing ReachBot to provide access to previously inaccessible space environments.
The main contributions of this work are threefold:
(1) We introduce the configurable parameters for ReachBot, a robot concept that uses extendable booms for mobility.
(2) We present a method to quantify the tradeoffs of a design by considering the robot configuration, terrain parameters, and mission requirements.
(3) We present a case study wherein we design ReachBot's configuration for a mission to a martian lava tube.

\textit{Paper Organization:} 
The rest of the paper is organized as follows. In Section~\ref{sec:related}, we discuss existing trade studies for mobile robot design and introduce a portfolio of past work in dexterous manipulation that can be applied to ReachBot. In Section~\ref{sec:methods}, we describe the workflow of our design process, including relevant parameters of the robot, terrain, and mission. A highlight of this section is building a relationship between ReachBot's boom configuration and various performance metrics. Then, in Section~\ref{sec:results}, we apply this design process to a case study exploring a martian lava tube.
Finally, we provide conclusions on the completeness of this process and how it might be applied to missions distinct from the case study where ReachBot would also excel.

%%%%%%%%%%%%%%%%%%%%%%%%%%%%%%%%%%%%%%%%%%%%%%%%%%%%%%%%%%%%%%%%%%%%%%%%%%%%%%%%
\section{Related work} \label{sec:related}
%%%%%%%%%%%%%%%%%%%%%%%%%%%%%%%%%%%%%%%%%%%%%%%%%%%%%%%%%%%%%%%%%%%%%%%%%%%%%%%%
Trade studies are widely used to develop robot designs that maximize certain performance metrics by intelligently selecting the robots' configurations in relation to their operational design domains.
For mobile robots, trade studies encompass a wide range of parameters from high-level mobility mode~\cite{ono2018enceladus} down to joint orientation~\cite{TedeschiCarbone2014}.
A valuable approach in designing a mission-specific robot is to relate environmental, task, and configuration parameters to performance metrics, then prioritize the performance metrics to align with mission objectives~\cite{apostolopoulos2001analytical}.

While many existing trade studies for mobile robots consider ground-based navigation metrics, such as a robot's ability to traverse soft soils or hard ground without loss of traction (trafficability),
designing ReachBot % a robot that relies on anchored mobility 
requires extra attention to performance criteria involving objects suspended by potentially long limbs.
As in past work, we approach this problem by leveraging ReachBot's similarities to a dexterous manipulator: whereas a manipulator hand uses its fingers to push on an object, ReachBot uses its booms to pull on the environment.
From this perspective, we consider metrics common in manipulator design that are relevant to ReachBot's performance such as stability, kinematic workspace, and manipulability~\cite{borras2015dimensional,ciocarlie2013kinetic,salisbury1982articulated}.

The stability of a grasped object is defined by the maximum external wrench (forces and torques) the grasp can resist in its weakest direction. The eigenvalues of the grasp stiffness matrix govern the grasp's ability to apply or resist wrenches, so its stability is determined by the minimum eigenvalue, where the corresponding eigenvector denotes the weakest direction. A grasp stiffness matrix with at least one negative eigenvalue describes a grasp that is unstable with no external applied wrench~\cite{howard1996stability}. Conversely, a grasp's wrench capability, i.e. the maximum wrench it can apply, is defined by the maximum eigenvalue of the grasp stiffness matrix. The grasp can apply this wrench in the direction of the corresponding eigenvector.

The kinematic workspace of a grasp defines bounds of the grasped object's position and orientation corresponding to a kinematically feasible configuration of fingers that does not exceed joint limits~\cite{borras2015dimensional}.
Within this workspace, manipulability is another measurement of the quality of a grasp. Manipulability of a grasp can be quantified by the size of the manipulability ellipsoid, which corresponds to the distance from singularities (points where the grasped object's movement is limited)~\cite{vahrenkamp2012manipulability}.
With multiple arms sharing the same workspace, mechanical interference must be taken into consideration as well~\cite{wahlvalidation}. %and also per arm contribution within the given workspace.
Our approach in this paper combines existing trade study methods for mobile robot design with design optimization techniques for manipulators to optimize ReachBot's configuration for its mission.

%%%%%%%%%%%%%%%%%%%%%%%%%%%%%%%%%%%%%%%%%%%%%%%%%%%%%%%%%%%%%%%%%%%%%%%%%%%%%%%%
\section{Trade study method}\label{sec:methods}
\begin{figure}[h]
    \centering
    \includegraphics[width=\columnwidth]{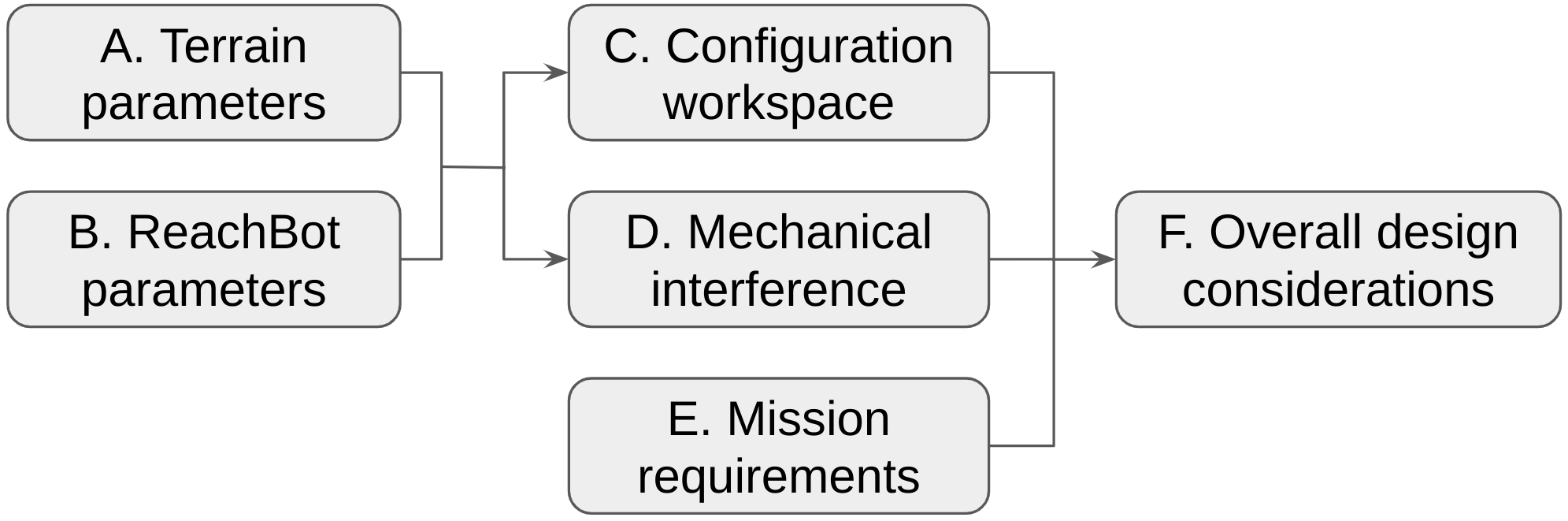}
    \caption{Designing ReachBot for a given mission proceeds from the interaction between the robot and terrain, then incorporates mission objectives to evaluate performance and deliver a final design.}
    \label{fig:flowchart}
\end{figure}

%%%%%%%%%%%%%%%%%%%%%%%%%%%%%%%%%%%%%%%%%%%%%%%%%%%%%%%%%%%%%%%%%%%%%%%%%%%%%%%%
In this section, we detail each step of our design process and define both configurable and fixed parameters. Fig.~\ref{fig:flowchart} shows the process overview for designing ReachBot in known terrain. First, we analyze the workspace and mechanical interactions of the robot and terrain. Then, incorporating mission requirements, we compute quantitative performance metrics and finalize the design. Because the exact terrain details may not be known during the design process, we use a Monte Carlo method to randomize terrain parameters, then design ReachBot to perform well over the portfolio of possible terrains. In this way, our design process focuses on maintaining versatility in unknown environments.

Section~\ref{sec:methods}A describes basis geometries and properties of the environment that affect how ReachBot interacts with it. Section~\ref{sec:methods}B identifies configurable parameters of ReachBot's design and asserts fixed values for a subset of those parameters to make the design process tractable.
Then, for a given set of terrain and robot parameters, the method presented in Section~\ref{sec:methods}C calculates the stability and manipulability of the configuration workspace. In Section~\ref{sec:methods}D, we present an approach to determine the extent of any mechanical interference between ReachBot's booms.
The workspace defines what the robot \textit{can} access, and the objectives of the mission, described in Section~\ref{sec:methods}E, define what the robot \textit{needs to} access. The overlap of these two areas determines how successful that robot will be at carrying out the mission, which we discuss in Section~\ref{sec:methods}F.

\subsection{A. Terrain parameters}
\begin{figure}
    \centering
    \includegraphics[width=\columnwidth]{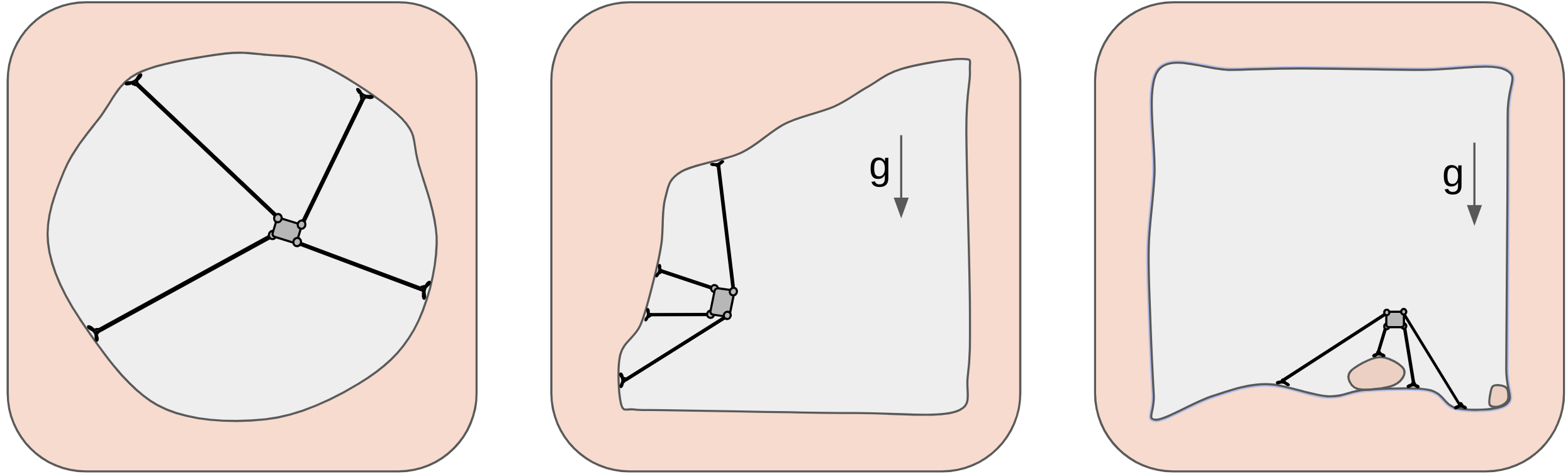}
    \caption{Three representative topoligies of different environments. In a corridor (left), ReachBot spans the width of the cavity and can act as a cable-driven robot. Against a wall (center) or floor (right), ReachBot interacts with a flat surface with operational modes differing because gravity is adversarial or cooperative, respectively.}
    \label{fig:topologies}
\end{figure}

Terrain parameters include the geometry and properties of the surface that affect ReachBot's operation.
For example, the different topologies illustrated in Fig.~\ref{fig:topologies} represent environments that might be encountered in different mission contexts. These topologies include: 

\begin{itemize}[leftmargin=1em]
    \item A corridor, which describes the terrain in a cave or lava tube where the span of the cavity is smaller than the span of the robot. A corridor does not require gravity for mobility, but it does require anchor points to which the end-effectors can attach. ReachBot spreads its booms out in all directions to achieve force closure.
    \item A wall or ceiling, which describes the terrain of a cliff with vertical or overhanging surfaces. A wall with anchor points does not require gravity for mobility, and in fact can be traversed when gravity is oriented to pull the robot away from the surface.
    \item A floor is flat ground where non-zero gravity is pushing the robot into the ground. Instead of grasping onto traditional anchor points, a robot would act as a legged walker, pushing off the ground to take steps.
\end{itemize}

Other terrain features include ungraspable smooth regions, surface irregularities like crevasses, and obstacles like rubble piles or boulders.
The size and distribution of these features determine whether ReachBot can step over them, climb onto them, or avoid them entirely. As the location and size of terrain features will be unknown to ReachBot a priori, our analysis randomizes anchor point availability to simulate varied terrain.
Lastly, the type of end-effector needed to traverse a surface depends on the surface material.

\subsection{B. ReachBot parameters}
\begin{figure}[b]
    \centering
    \includegraphics[width=\columnwidth]{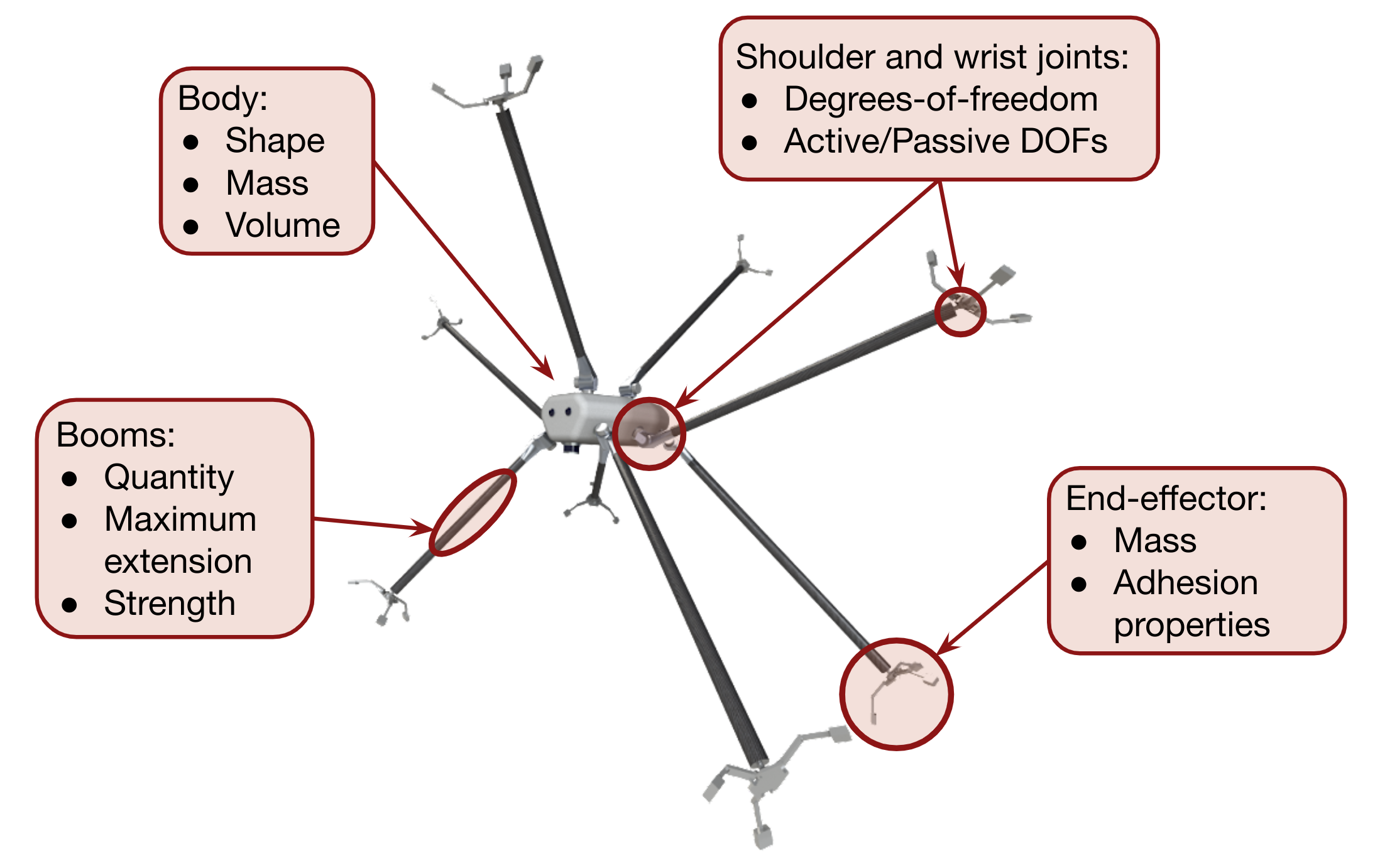}
    \caption{ReachBot's configurable parameters include the size of the body, design of rotational (shoulder and wrist) joints, boom quality and material properties, and end-effector design}
    \label{fig:system_design}
\end{figure}

This trade study considers four major areas of configurable parameters: (1) body size, (2) rotational joints, (3) booms, and (4) end-effectors. These areas are highlighted in Fig.~\ref{fig:system_design}.
Unlike in our past work where we assumed a planar rectangle with four booms, in this trade study we vary ReachBot's configurable parameters and observe how these changes affect performance to determine an optimal architecture.

Before making mission-specific design decisions, we assign values to some of these parameters to make the trade study tractable. 
First, we use fundamental properties of deployable booms to constrain actuation capabilities. Deployable booms are known to have significantly stronger tensile and compression strength than bending~\cite{Opterus2022}, so ReachBot's design should discourage applying bending moments.
While the shoulder must have active 2-degree-of-freedom (DoF) control (e.g., with a pan-tilt joint) to aim the end-effector to a specific target, a back-drivable motor allows the shoulder to act like a passive joint when the end-effector is attached, minimizing the bending moment applied to the boom from the shoulder. Similarly, a passive wrist joint prohibits the wrist from applying a bending moment while attached \textit{and} reduces the weight of the end-effector assembly, which lessens boom bending while detached. The passive wrist also allows the gripper to conform to the orientation of the target grasp point. So, for all configurations of ReachBot, we assume each boom will be mounted on a back-drivable pan-tilt shoulder joint with $\pi/4$ rad range-of-motion in each direction, as per industry standards. Additionally, we assume the gripper is mounted on a spring-loaded passive ball joint at the wrist.

The structural properties of ReachBot's booms are crucial to its success, but current boom designs have been optimized for on-orbit operation and existing technology is unsatisfactory for planetary use.
However, the strength and stiffness of deployable booms follow predictable scaling laws and in general are not limited to theoretical capabilities~\cite{Opterus2022}.
Instead of incorporating specific boom selection into this trade study, we continue our analysis under the assumption that when needed, booms will be manufactured to meet ReachBot's structural requirements with proportional cost, mass, and volume.
An initial analysis on boom cross-section design demonstrates the growing potential of deployable boom technology: the newer collapsible tube mast booms, also known as lenticular or double omega, have favorable buckling properties as compared to traditional slit-tube or triangular booms~\cite{LeclercPellegrino2020, francis2015high, stohlman2021characterization}.

Finally, we rely on state-of-the-art grippers for a given terrain surface to characterize ReachBot's end-effectors.
State-of-the-art designs for grasping onto rock, ice, or smooth surfaces specify predicted adhesion properties, weights, and operational modes. While ReachBot is designed to be end-effector agnostic, the specifics of its end-effectors must be incorporated into overall design decisions.
For example, for rocky missions, we will consider end-effectors with the properties of our microspine grippers currently under development for grasping the walls of lava tubes~\cite{ChenMillerEtAl2022}. For icy missions, we look to existing technology in icy grippers, for example ice screw end-effectors~\cite{curtis2018roving}.
These assertions narrow the design space, but leave many of ReachBot's parameters to be optimized for the mission.

\subsection{C. Configuration workspace}
To compute stability and manipulability of ReachBot's workspace, we use terrain and robot parameters to represent ReachBot as a manipulator.
While some robot architectures are designed for a specific terrain, versatility is a higher priority in unpredictable environments. 
Therefore, we assess stability and manipulability of a workspace defined in relation to randomized terrain features to analyze ReachBot's configuration independently of (unknown) terrain.

% grasp stability
ReachBot's stability is defined as its ability to resist any external wrench, i.e., to remain stable in the presence of an applied force and/or torque.
For a given assignment of end-effectors to available anchor points, we calculate a grasp stiffness matrix as if ReachBot's booms were manipulator fingers ``holding" the robot body. Then, we find the smallest eigenvalue of that matrix to determine the maximum magnitude wrench ReachBot can resist independent of direction.
Conversely, the maximum eigenvalue of the grasp stiffness matrix corresponds to wrench capability, or the maximum magnitude wrench that ReachBot can apply in its strongest direction.

\begin{figure}[tp]
    \centering
	\subfloat[ReachBot stance with three booms]{
    \begin{subfigure}[]{0.4\columnwidth}
	\includegraphics[width=\textwidth]{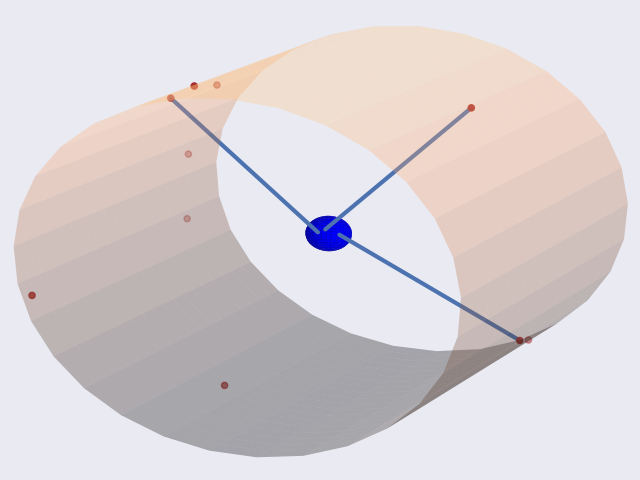}
    \end{subfigure}
    \hspace{4pt}
    \begin{subfigure}[]{0.4\columnwidth}
	    \includegraphics[width=\textwidth]{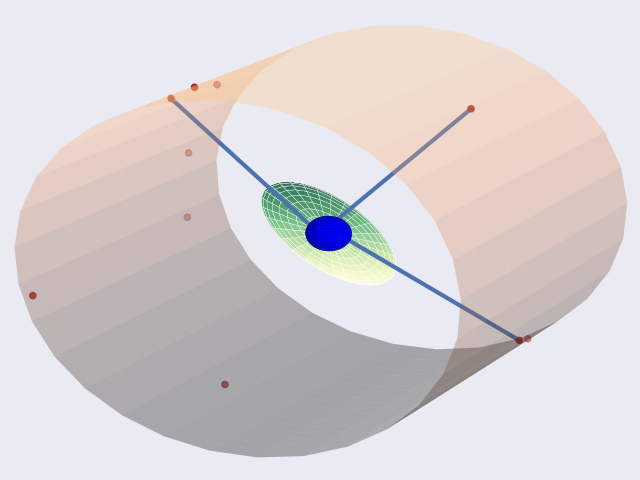}
	\end{subfigure}
    }
    
	\subfloat[ReachBot stance with six booms]{
    \begin{subfigure}[]{0.4\columnwidth}
	\includegraphics[width=\textwidth]{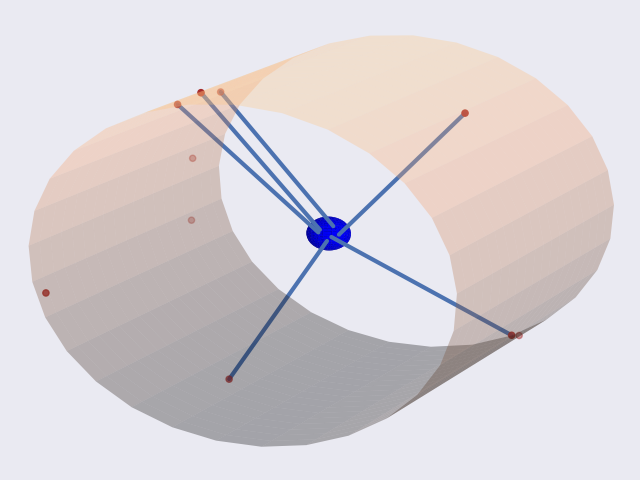}
    \end{subfigure}
    \hspace{4pt}
    \begin{subfigure}[]{0.4\columnwidth}
	    \includegraphics[width=\textwidth]{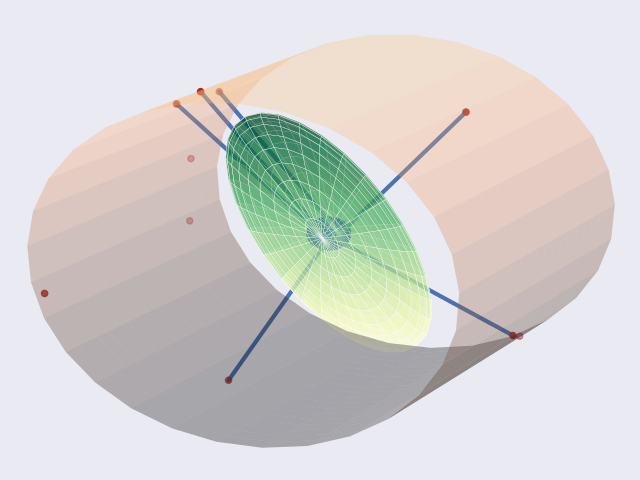}
	\end{subfigure}
    }
    
    \caption{Two different ReachBot configurations in a simplified cylindrical topology: one each with three and six booms. The topology is represented as a cylinder with randomized anchor points. ReachBot's stance is shown without (left) and with (right) manipulability ellipse overlaid. Only the positional manipulability ellipsoid is visualized, but the full 6D ellipsoid is considered in the trade study.}
     \label{fig:manipulability}
\end{figure}

Since the anchor point availability is unknown a priori, we evaluate a configuration's stability by randomizing anchor point locations within the defined topology. To illustrate this approach, consider the cylinder in Fig.~\ref{fig:manipulability}, which is one possible 3D abstraction of a corridor topology.
Let ReachBot be a sphere placed at the center of the cylinder with $N$ booms equally spaced on its body. We generate $M \ge N$ uniformly random anchor points, then we assign each of ReachBot's $N$ booms to an anchor point such that the assignment minimizes total deployed boom length. This heuristic discourages stances that abut joint limits.
The minimum eigenvalue of the stiffness matrix quantifies ReachBot's stability in each stance of randomly selected anchor points. By considering multiple instantiations of anchor points, we get a sense of the configuration's stability in terrain with the specified geometry but unknown features.

% manipulability
Manipulability of a grasp is a measurement of a hand's ability to move the grasped object within a workspace, i.e., to avoid singularities.
% It is measured with respect to a kinematic workspace, which bounds the grasped object's position and orientation. ReachBot's kinematic workspace corresponds to the positions and orientations the body can assume given an assignment of end-effectors to anchor points. To calculate the kinematic workspace, we systematically consider every position in the convex hull of the anchor points and solve inverse kinematics to determine if that pose is feasible within joint limits. This results in a volume of kinematically feasible poses. Then, we compute the manipulability ellipsoid at every point within ReachBot's kinematic workspace.  Fig.~\ref{fig:manipulability} illustrates the manipulability ellipsoid at various points in the workspace, where red and blue arrows signify the principle axes of larger and smaller ellipsoids, respectively. A larger manipulation ellipsoid corresponds to higher manipulation authority, i.e., a higher capacity to change ReachBot's position and orientation from a given joint configuration. To dissociate configuration quality with a specific anchor point geometry, we perform this computation for several instances of randomized anchor points.
Many common measures of grasp manipulability involve analyzing the manipulability ellipsoid that spans the singular vectors of the grasp Jacobian
%, for example, measuring its volume or the relation of the smallest to largest singular value
~\cite{yoshikawa1985manipulability,togai1986application}.
For ReachBot, the manipulability ellipsoid signifies the position and orientation changes ReachBot can enact within its stance.
While the manipulability ellipsoid can be optimized for a particular task~\cite{vahrenkamp2012manipulability}, we use a more general metric
\begin{equation}
    w = \sqrt{\det (J J^\top)},
    \label{eq:w}
\end{equation}
where $J$ is the grasp Jacobian and $w$ is proportional to the volume of the manipulability ellipsoid. By using this metric rather than task-specific criteria, we preserve ReachBot's versatility in unknown terrain.
A more voluminous manipulability ellipsoid corresponds to higher manipulation authority.
Fig.~\ref{fig:manipulability} illustrates the positional manipulability ellipsoid for two ReachBot configurations (with three and six booms) in a cylindrical geometry with the same set of randomized anchor points. 
To dissociate configuration quality with a specific anchor point placement, we calculate the manipulability of a configuration by computing $w$ for several instances of randomized anchor points.

\subsection{D. Mechanical Interference}
When considering stability and manipulability, it is beneficial to add more booms or bigger actuators to increase performance. However, an important advantage of ReachBot over articulated-arm robots is simplicity, namely from concentrating the actuation of each limb at a single joint, but we compromise this simplicity by adding more booms. Therefore, we must consider the mechanical interference between booms to ensure the benefits of more booms outweigh the costs of added complexity. 

\begin{figure}[bp]
    \centering
    \begin{subfigure}[]{0.4\columnwidth}
	\includegraphics[width=\textwidth]{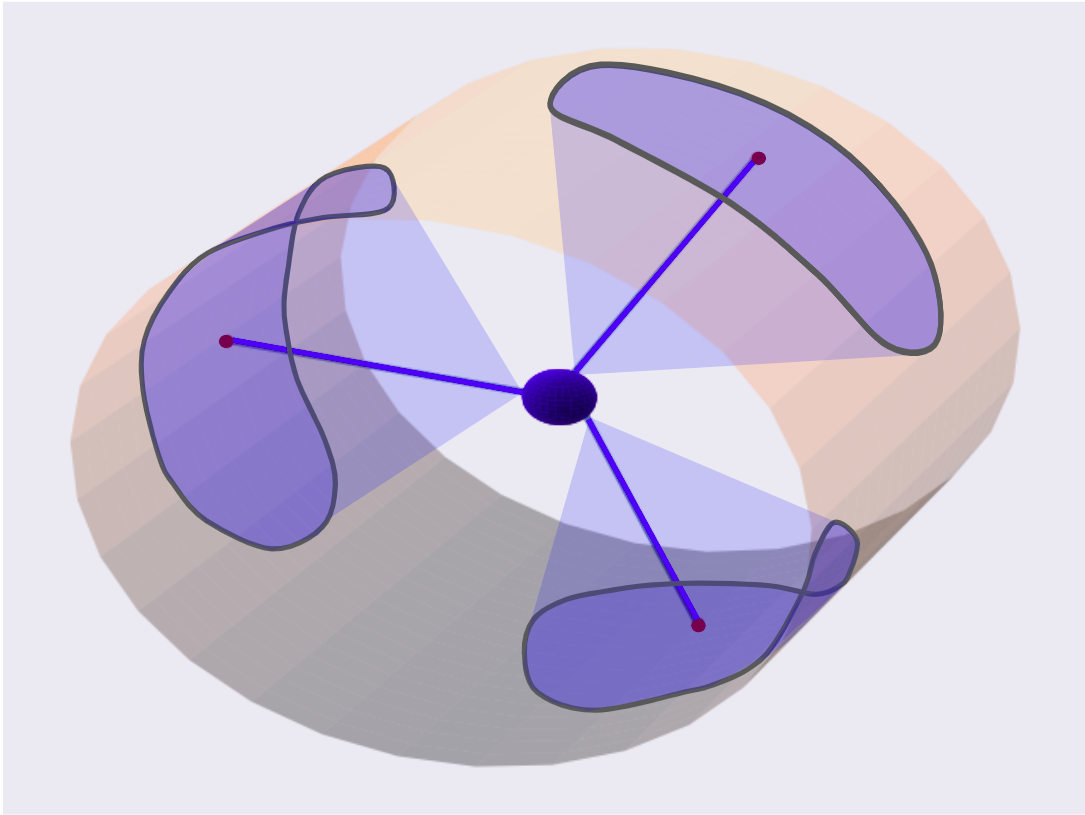}
    \end{subfigure}
    \hspace{4pt}
    \begin{subfigure}[]{0.4\columnwidth}
	    \includegraphics[width=\textwidth]{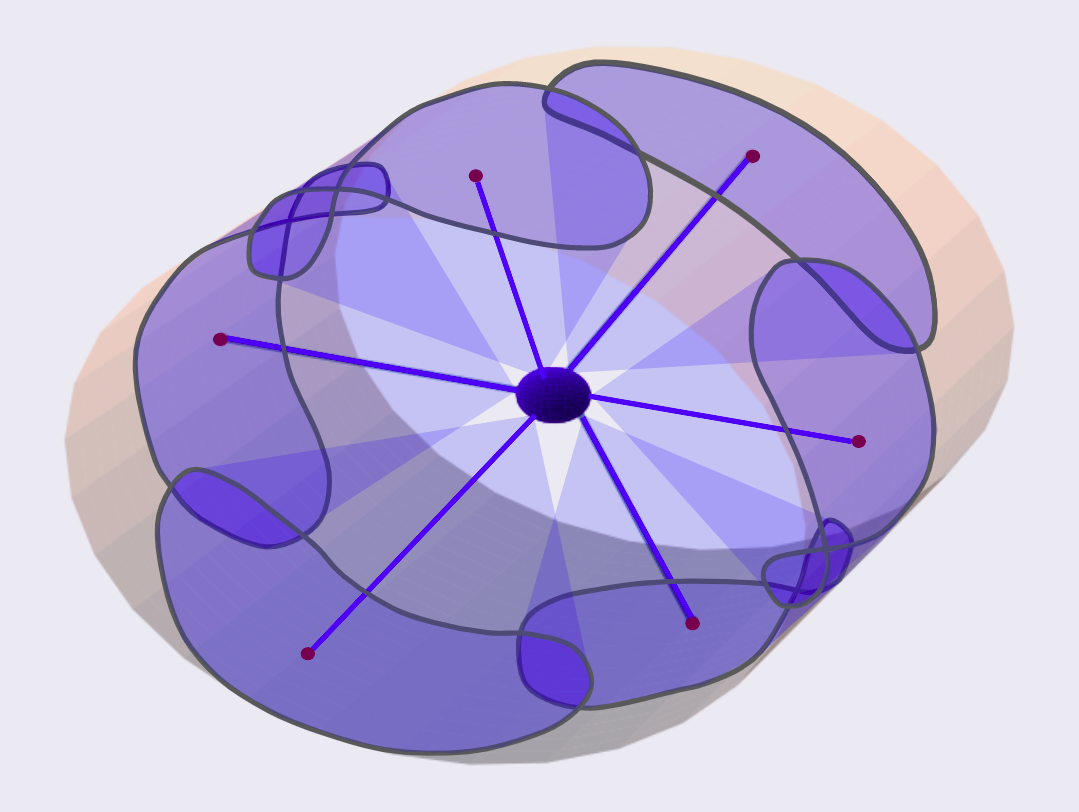}
	\end{subfigure}

    \caption{Two different ReachBot configurations in a simplified cylindrical topology: one with three booms (left) and one with six booms (right). For each configuration, the booms are spaced uniformly to match the topology. The range-of-motion of each boom is the same, but with three booms, ReachBot can access only a small fraction of the terrain. With six booms, ReachBot can access significantly more of the terrain, but the accessible regions (shown as elliptical intersections of cones with the cylinder) overlap.}
     \label{fig:mech}
\end{figure}

To quantify interference of a configuration, we look at the full range-of-motion of each boom and determine the surface area it can access, which directly translates to the number of anchor points it can reach. To assess the value of one boom, we consider not only how much area it can reach, but how much of that area can be reached by other booms. In other words, we calculate the per-boom contribution to expanding the accessible workspace.
%To quantify interference of a configuration, we look at the full range-of-motion of each boom and assess not only how many anchor points are accessible, but also any overlapping accessible regions of two or more booms.
While the booms can be controlled to avoid self-collision in these overlapping regions, having multiple booms able to access the same anchor points indicates unnecessary redundancy.

To illustrate our method of quantifying mechanical interference, consider the cylindrical terrain in Fig.~\ref{fig:mech}. Each boom's range-of-motion is represented as a cone, and the intersection of that cone with the cylinder encompasses that boom's accessible terrain. With only three booms, ReachBot can only reach a small fraction of the terrain, but each boom provides access to previously inaccessible terrain. With six booms, ReachBot can access significantly more of the terrain, but the accessible regions overlap. Increasing the number of booms increases accessible surface area, but with diminishing value as new booms add access to terrain that was already accessible.

\subsection{E. Mission requirements}
Mission requirements bridge the gap between an isolated prototype and a comprehensive system design. These requirements come from environmental factors, such as temperature, pressure, and dust, as well as operational details like the inclusion of a tether or mother craft. 
Requirements like payload capacity, traversability, and manipulation capabilities are derived from the mission objective. %They may either provide constraints on the final design or simply encourage behaviors
For example, a mission searching for signs of astrobiology may require a robot to take measurements with centimeter-scale accuracy, providing constraints for ReachBot's task-specific manipulability ellipsoid.
For another example, a sample-retrieval mission requires a robot capable of applying a large wrench to drill and extract samples, providing necessary lower bounds on wrench capability and stability.

\subsection{F. Overall design considerations}
Our design process produces quantitative metrics such as stability, manipulability, and mechanical interference that are presented as functions of boom configuration. Other metrics such as cost, mass, and stowed volume tend to increase linearly with additional booms.
One other consideration is the reliability of the system as a function of boom configuration.  On one hand, additional booms increase the stability, manipulability, and redundancy of the robot, but the additional actuators and other components increase the number of potential failure points. The design should consider this balance between redundancy and risk of failure.

To compare the performance of different configurations, we need to balance multiple objectives by weighing them according to mission priority. For example, a ReachBot design suited for a mission with high manipulability and workspace requirements might be incompatible with a mission where ReachBot is a secondary payload with a strict mass budget.
From the objectives that are not limited by hard constraints, we select a Pareto-optimal design that best matches the given mission.

%%%%%%%%%%%%%%%%%%%%%%%%%%%%%%%%%%%%%%%%%%%%%%%%%%%%%%%%%%%%%%%%%%%%%%%%%%%%%%%%
\section{Case study: martian lava tube exploration}\label{sec:results}
%%%%%%%%%%%%%%%%%%%%%%%%%%%%%%%%%%%%%%%%%%%%%%%%%%%%%%%%%%%%%%%%%%%%%%%%%%%%%%%%
In this section, we step through our design process to generate a ReachBot configuration tailored to a mission to a martian lava tube.
Lava tubes and other cavernous features on Mars that would present ancient materials have recently
been identified as promising locations of astrobiological interest. Due to their insulating and
shielding properties, these caverns provide relatively stable conditions which may promote
mineral precipitation and microbial growth, as well as useful sites for future human habitation~\cite{LeveilleDatta2010,Boston2010}. To maximize scientific value, a robotic mission to a martian lava tube should take precise measurements, drill, and gather samples from a wide range of strata and locations along the interior of the tube.

\subsection{A. Terrain parameters}
\begin{figure}[t]
    \centering
	\subfloat[Side view of representative martian lava tube]{
		\label{fig:mars_side}
	\includegraphics[width=\columnwidth]{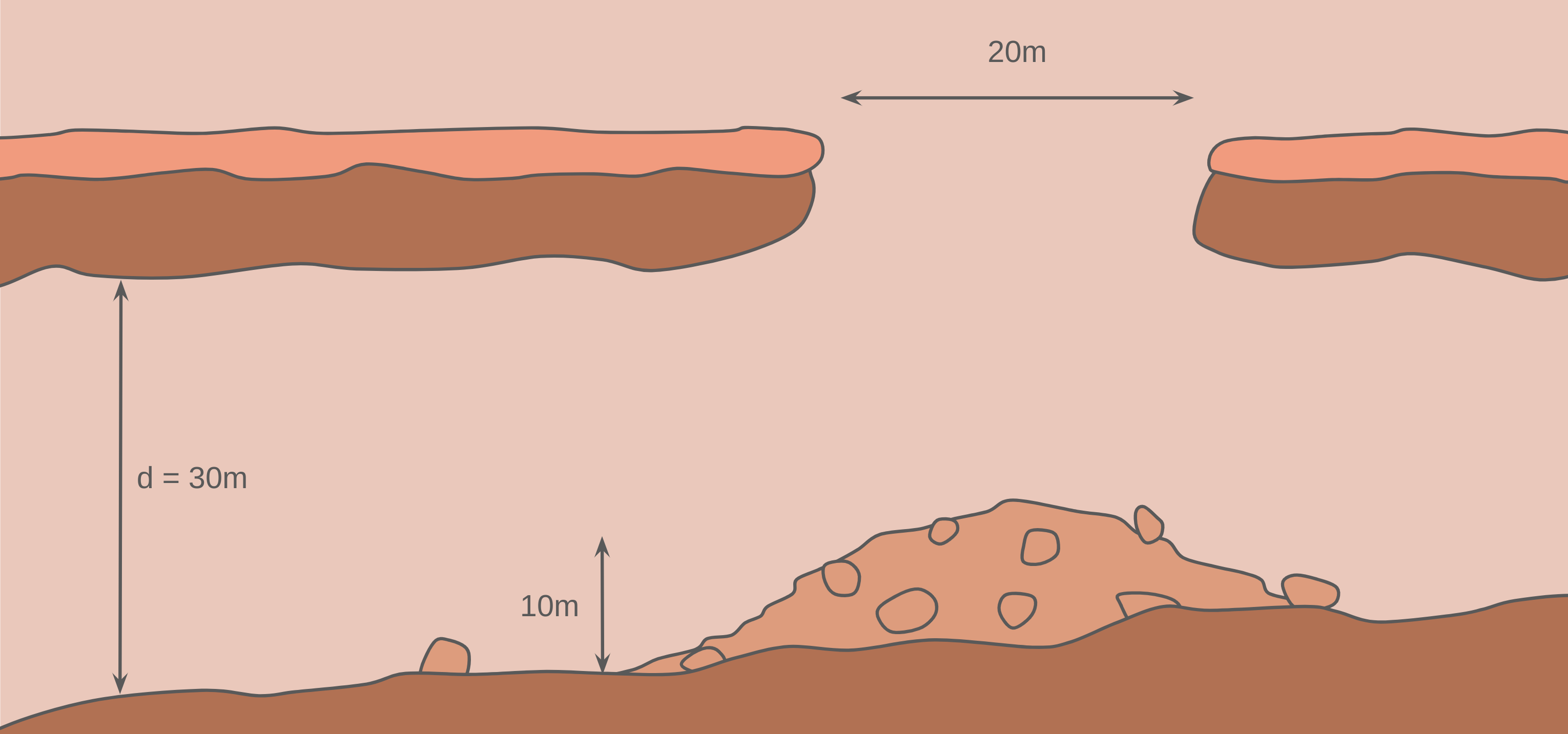}
    }
    
	\subfloat[Front view of representative martian lava tube]{
		\label{fig:mars_front}
	\includegraphics[width=\columnwidth]{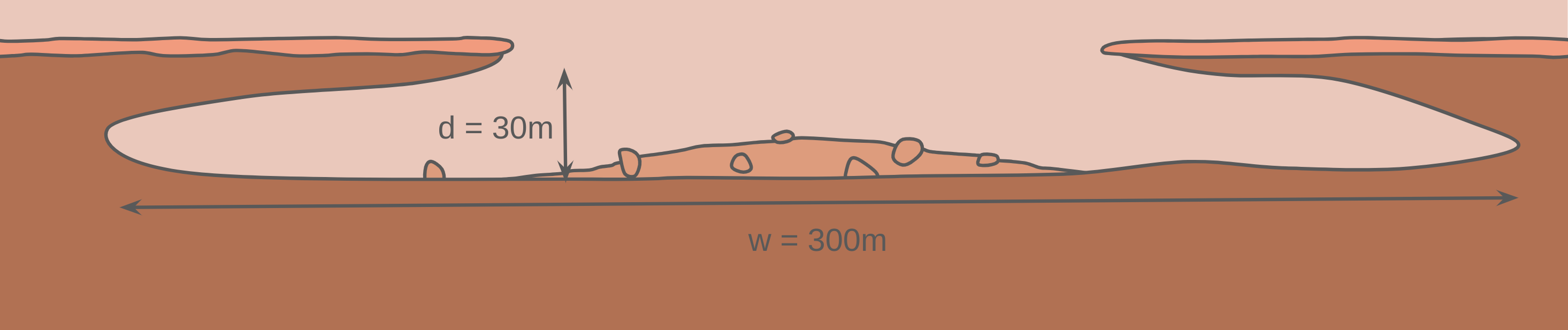}
    }
    \caption{Illustration of a martian lava tube with an average depth of $30$m and width of $300$m. The example shown here has a debris pile underneath a skylight, both of which are features believed to be present in several lava tubes on Mars.}
    \vspace{-10pt}
     \label{fig:mars_lava_tube}
\end{figure}

A representative lava tube environment is illustrated in Fig.~\ref{fig:mars_lava_tube}. We constructed an example lava tube based on average estimated dimensions for depth (30m) and width (300m)~\cite{SauroPozzobonEtAl2020}. This fictional environment is consistent with studies that suggest boulders, rockfall, and other debris may form mounds under skylights~\cite{WilkensIliffeEtAl2009}.

\subsection{B. ReachBot parameters}
Using the constraints imposed by the terrain, we begin to prescribe ReachBot's configuration parameters. 
We notionally assign ReachBot a mass of $10$kg, which aligns with estimates from a similar class robots~\cite{KennedyOkonEtAl2006,Parness2017}.
%To meet the mission's sample-return objective, we assume a payload of equal size and mass to the turret-mounted corer included in the sample acquisition system onboard Perseverance, which weights $23.5$kg and can be approximated as a 80cm x 80cm x 100cm rectangular prism~\cite{moeller2021sampling}.
To leverage force closure, we define a notional maximum boom extension of $20$m so ReachBot can span the depth of the lava tube with extra range for mobility around obstacles.
Additionally, we equip ReachBot with lightweight microspine grippers that have a high adhesion force to weight ratio, for example those under current development for our NIAC project~\cite{PavoneCutkoskyEtAl2022}.
The adhesion from only 2-3 of these grippers supports the weight of the robot on Mars, so ReachBot can assume a wide stance as in the corridor in Fig.~\ref{fig:topologies} and keep its deployable booms in tension. In this operational mode, the booms will not be loaded in compression, so the limiting case for boom failure is buckling. Specifically, the highest buckling load arises when a boom is fully outstretched. In this configuration,
the boom must support the weight of the gripper at the end of a fully deployed boom and the weight of the boom itself. This moment is given by
\begin{equation}
    M_\text{shoulder} = m_\text{gripper}\: g_\text{Mars}\: L + \frac{1}{2}\: m_\text{boom}\: g_\text{Mars}\: L,
\end{equation}
where $M$ is the buckling moment felt at the shoulder, $m$ is mass, $g_\text{Mars} = 3.721\text{m}/\text{s}^2$, and $L = 20$m is the length of the fully deployed boom. Publicly available boom designs would be prohibitively large to support their own weight under martian gravity. However, active development that includes confirmed fabrication of a $20$m boom by Opterus suggests upcoming improvements in boom structural properties and manufacturing technology. Therefore, we assume a mass of $1$kg per boom and leave it as a requirement that the critical buckling strength $M_{CR}$ satisfies the constraint
\begin{equation}
    M_{CR} > M_\text{shoulder},
\end{equation}
deferring the specifics of boom design to future work. The following sections justify each boom by balancing mass and volume costs with the benefits of more booms.

\subsection{C. Configuration workspace}
\begin{figure}[bp]
    \centering
	\subfloat[Stability, denoted by maximum wrench resistance, for up to 10 booms calculated for 100 grasps with randomized anchor point placement.]{
		\label{fig:stab_results}
	\includegraphics[width=\columnwidth]{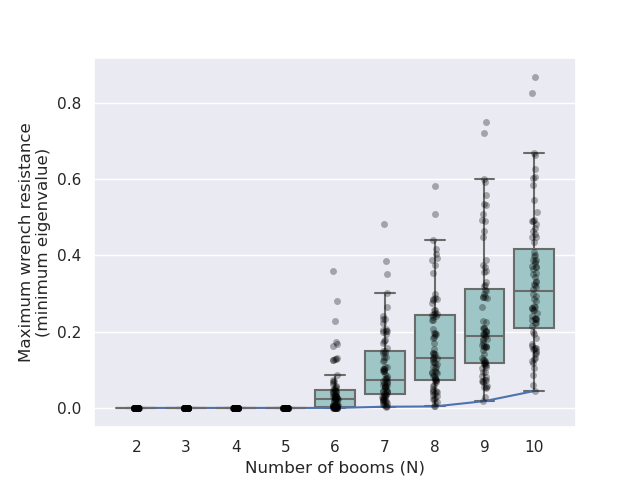}
    }
    \\
	\subfloat[Average stability increase from adding one boom.]{
		\label{fig:stab_increase}
	\includegraphics[width=\columnwidth]{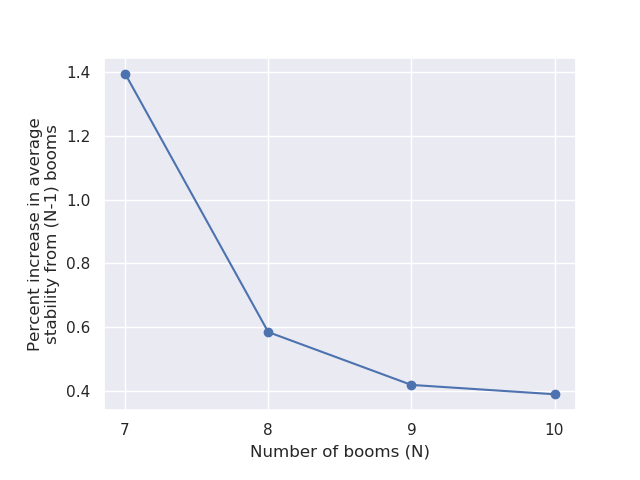}
    }
    \caption{Stability of ReachBot with a variable number of booms. Stability is quantified by the minimum eigenvalue of the grasp stiffness matrix, which represents the maximum wrench a grasp can resist in its weakest direction.}
        \vspace{-10pt}
     \label{fig:stability}
\end{figure}

By employing the methods
described in Section~\ref{sec:methods}C, we achieve estimates of ReachBot's stability and manipulability in a cylindrical ``corridor" terrain. 
Stability, given by the minimum eigenvalue of the grasp stiffness matrices evaluated over 100 Monte Carlo trials of randomized anchor point placement (held consistent across different ReachBot configurations), is shown in Fig.~\ref{fig:stab_results}. The minimum stability measure across all trials for a given ReachBot configuration is shown as a blue line, representing the worst-case stability for a given number of booms. As expected, any fewer than six booms creates a 6D system that cannot resist an arbitrary wrench.
Although ReachBot's stability monotonically increases with the number of booms, the diminishing returns seen in Fig.~\ref{fig:stab_increase} suggest that there exists an optimal number of booms for which the stability gained by adding more is not worth the additional cost.

\begin{figure}[bp]
    \centering
    \includegraphics[width=\columnwidth]{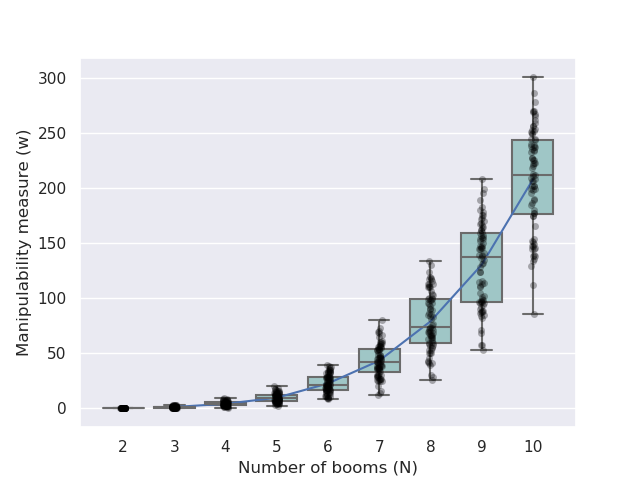}
    \caption{Manipulability of a configuration is quantified by $w$ from~\eqref{eq:w} which is proportional to the volume of the manipulability ellipsoid, calculated for 100 grasps with randomized anchor point placement. 
    }
    \label{fig:man_results}
\end{figure}

Manipulability over 100 Monte Carlo trials of randomized anchor point placement is shown in Fig.~\ref{fig:man_results}. 
While the volume of the manipulability ellipsoid increases superlinearly with the number of booms, we expect that more task-specific manipulability metrics which may take into account properties of the objects to be manipulated (e.g., desired force, maximum mass), will show diminishing returns as well.
For this reason, we don't weigh this metric of manipulability as highly in our final evaluation.
 
\subsection{D. Mechanical Interference}
\begin{figure}
    \centering
    \includegraphics[width=\columnwidth]{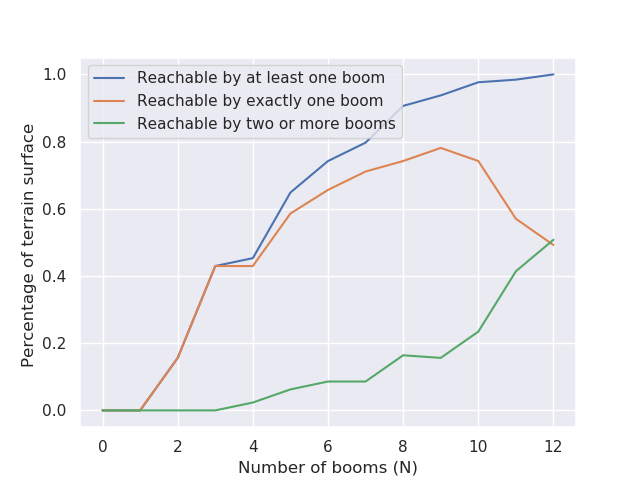}
    \caption{Percentage of surface area covered as a function of number of booms. Both unique coverage and overlapping coverage increase monotonically with more booms, but the benefits of additional booms diminishes after nine, when adding more booms adds more overlap than newly accessible terrain.
    }
    \label{fig:mech_results}
    \vspace{-10pt}
\end{figure}

To quantify the mechanical interference between ReachBot's booms, we consider surface area coverage as a function of the number of booms.
As seen in Fig.~\ref{fig:mech_results}, the percentage of accessible surfaces increases monotonically with more booms. However, the percentage of terrain with overlapping coverage increases at a higher rate.
While it is beneficial to access more of the terrain (higher surface area coverage),
there is little value in redundant access from multiple booms. Because of the growing overlap, we see diminishing returns on accessible terrain for any more than nine booms.

\subsection{E. Mission requirements}
% Butchered from NIAC Phase I Report
The scientific goals proposed in NASA's Decadal Survey~\cite{NRC2011} require a comprehensive in-situ study of martian caves and lava tubes that includes a combination of remote sensing and contact observations. While remote sensing enables overall mapping of walls and other geological features, contact observations, particularly of centimeter-scale targets, are key to characterizing geological, biological, and biochemical processes. These proximal observations
%are crucial for understanding the geological record and 
emphasize the importance of bringing scientific instruments, for example a microscopic imager or imaging spectrometer, adjacent to the surface~\cite{PavoneCutkoskyEtAl2022}.
ReachBot must therefore be equipped with an instrument suite that can be brought close to the surface without obstruction. Additionally, to gather samples, ReachBot must be able to apply enough torque to operate a drill.

To investigate relevant scientific targets and record context of measurements in deeper stratified layers, ReachBot needs to traverse rough and unpredictable terrain. Such navigation requires regular anchor point detection and constant scanning of the environment, particularly in unexplored areas. So, ReachBot should have a LiDAR or other perception system mounted toward the robot's direction of travel.

In addition to mobility challenges, subsurface exploration complicates other subsystem designs by occluding direct access to solar power and communication. We address the challenge of accessing power and communications by incorporating a tether for direct connection to the surface. In a cylindrical environment, ReachBot's booms should extend radially from its body to maximize versatility, but these mission requirements impede rotational symmetry by constraining the placement of shoulder assemblies on Reachbot's body. In particular, the booms must be placed to leave room for (1) an instrument suite that can get close to the surface, (2) a front-facing LiDAR assembly, and (3) a rear-facing tether spool. 

\subsection{F. Overall design considerations}
Now, we define the full system design and assess its suitability to the mission.
First, the design must meet our operational constraints.
The minimum bound on number of booms is determined by a requirement that ReachBot be stable in 6D even while it is suspended by one fewer boom during a footstep.
While stability represents the maximum wrench ReachBot can resist in its weakest direction, we also need to consider manipulation capability, which is the maximum wrench ReachBot can apply in its strongest direction.
To achieve the mission objective of gathering samples, ReachBot must be able to apply enough torque to drill while holding itself still. This task sets a lower bound on ReachBot's manipulation capability.
As a nominal bound, estimates for drilling into regolith suggest a required minimum applied torque of $4$Nm~\cite{hamade2010compact}.
%The size of the manipulability ellipsoid will be determined by expected spacing of anchor points, so we know that it will be able to move to a transition stance so it can take a step.

\begin{figure}
    \centering
    \includegraphics[width=\columnwidth]{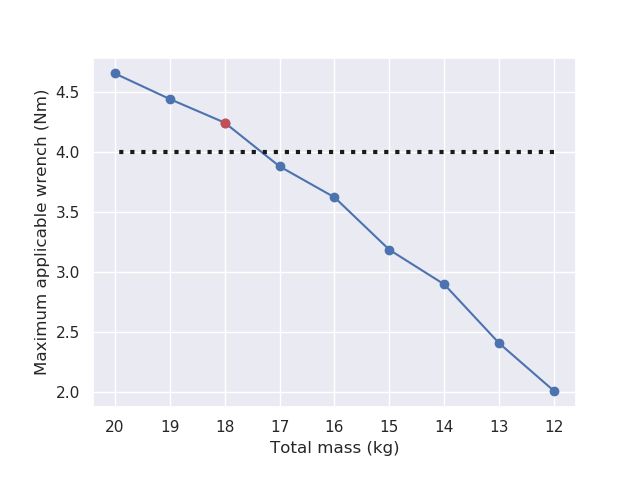}
    \caption{Two-dimensional slice of the Pareto front as a function of number of booms. The two objectives shown here are maximum applicable wrench and system mass, with a lower bound of $4$Nm on applied wrench to represent drilling requirements.
    }
    \label{fig:pareto}
\end{figure}

\begin{figure}
    \centering
    \includegraphics[width=\columnwidth]{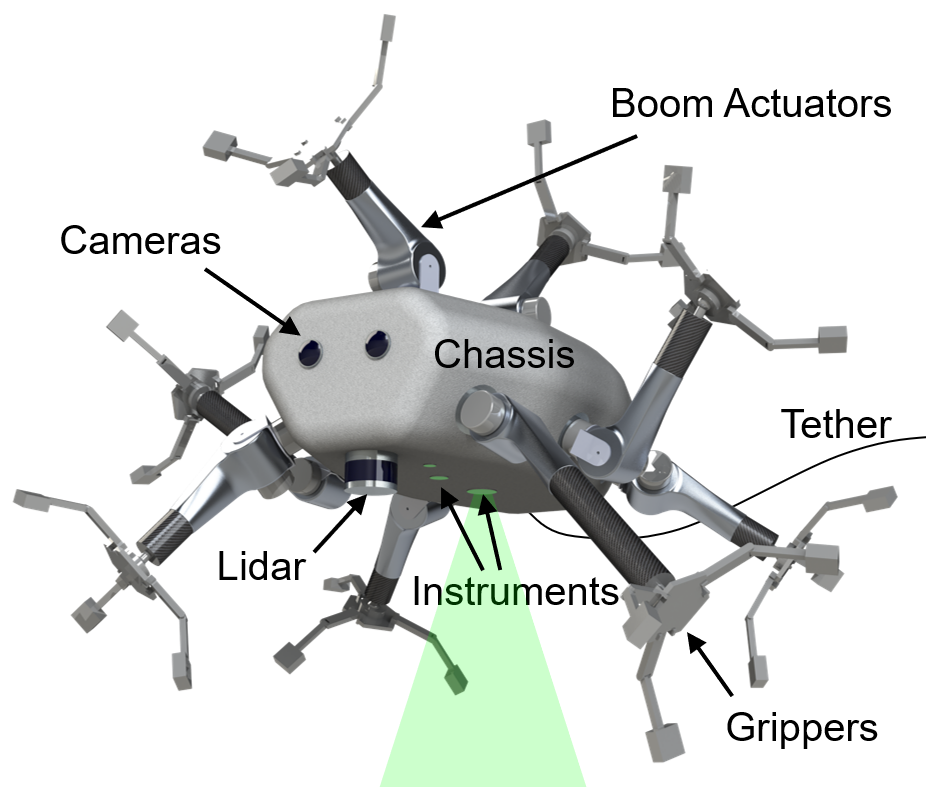}
    \caption{Final ReachBot configuration with labeled subsystems. The lateral and overhead boom positions leave room for cameras and LiDAR at the front, an instrument suite on the bottom, and a tether trailing out the back.
    }
    \label{fig:subsystems}
\end{figure}

To select the optimal number of booms while meeting these constraints, we consider the Pareto front of the design's multiple objectives. In the full design space, the Pareto front will be a high-dimensional surface, but for ease of visualisation, Fig.~\ref{fig:pareto} displays two objectives -- maximum wrench and system mass -- as a function of number of booms. Assuming ReachBot can position itself into an ideal drilling stance, its maximum applicable wrench is the maximum eigenvalue of the grasp stiffness matrix. The minimum bound on an applicable wrench is shown by the dotted line. For this mission, a design with eight booms satisfies stability and wrench capability constraints while minimizing mass and favoring lower mechanical interference. Fig.~\ref{fig:final_config} shows the final configuration of ReachBot traversing a lava tube, and 
Fig.~\ref{fig:subsystems} shows a close-up view of ReachBot with labeled subsystems. 

%%%%%%%%%%%%%%%%%%%%%%%%%%%%%%%%%%%%%%%%%%%%%%%%%%%%%%%%%%%%%%%%%%%%%%%%%%%%%%%%
\section{Conclusion}
%%%%%%%%%%%%%%%%%%%%%%%%%%%%%%%%%%%%%%%%%%%%%%%%%%%%%%%%%%%%%%%%%%%%%%%%%%%%%%%%
In this paper, we described a process to develop a mission-specific ReachBot configuration that incorporates terrain parameters and mission requirements into the design. We introduced methods to analyze ReachBot's stability, manipulability, and mechanical interference, and related those metrics to multiple competing mission objectives to determine the final configuration. We demonstrated this process for a mission to explore and collect samples in a martian lava tube. The final configuration for this mission has eight booms positioned around an oblong body with designated positioning of a leading LiDAR, trailing tether, and science suite on its underside.

In the future, we will apply this trade study approach to other representative missions with contrasting terrain parameters to exemplify ReachBot's capabilities. For example, a mission to descend a lunar pit crater leverages ReachBot's anchored climbing to stay close to overhanging surfaces. A mission to walk along the surface of an icy body like Enceladus uses ReachBot's booms as legs to traverse terrain that has obstacles of unknown size due to poor image resolution. The trade study presented here can be applied directly to these other missions by substituting in appropriate terrain parameters and mission requirements.

\vspace{-10pt}
\acknowledgments
Support for this work was provided by NASA under the Innovative Advanced Concepts program (NIAC) and by the Air Force under an STTR award with Altius Space Machines.
Stephanie Newdick is supported by a NASA NSTGRO fellowship.

%%%%%%%%%%%%%%%%%%%%%%%%%%%%%%%%%%%%%%%%%%%%%%%%%%%%%%%%%%%%%%%%%%%%%%%%%%%%%%%%
\bibliographystyle{IEEEtran} 
\bibliography{ASL_papers,main,new}

%%%%%%%%%%%%%%%%%%%%%%%%%%%%%%%%%%%%%%%%%%%%%%%%%%%%%%%%%%%%%%%%%%%%%%%%%%%%%%%%
\thebiography
\begin{biographywithpic}
{Stephanie Newdick}{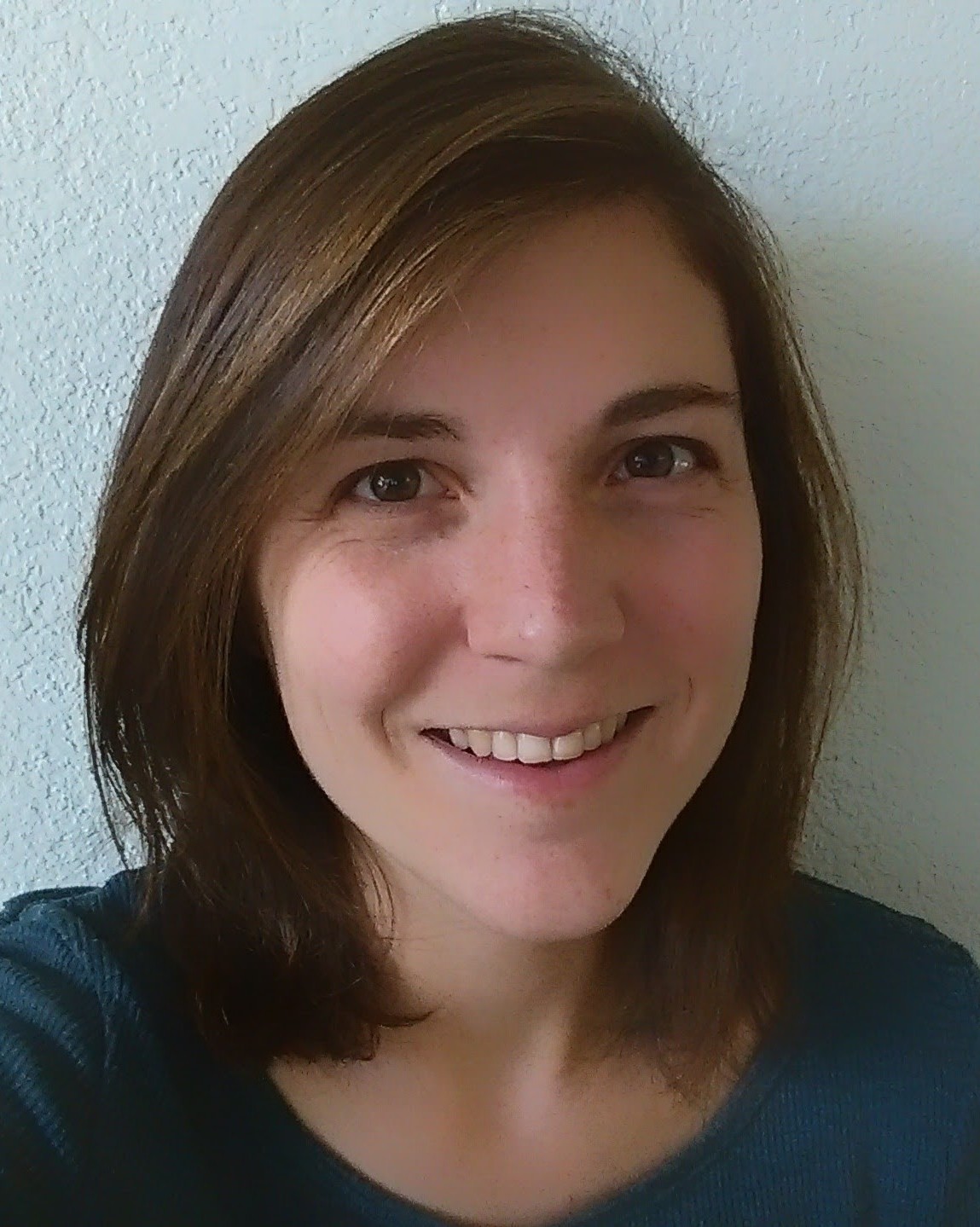}
is a Ph.D. candidate in the Autonomous Systems Lab in the Aeronautics and Astronautics Department at Stanford.
She received her B.S. in Mechanical Engineering from Cornell University in 2014. Prior to coming to Stanford, she worked as a software engineer and flight test engineer for Kitty Hawk Corporation. She is currently supported by an NSTGRO fellowship. Stephanie’s research interests include real-time spacecraft motion-planning, grasping and manipulation in space, and unconventional space robotics.
\end{biographywithpic}

\begin{biographywithpic}
{Tony G. Chen}{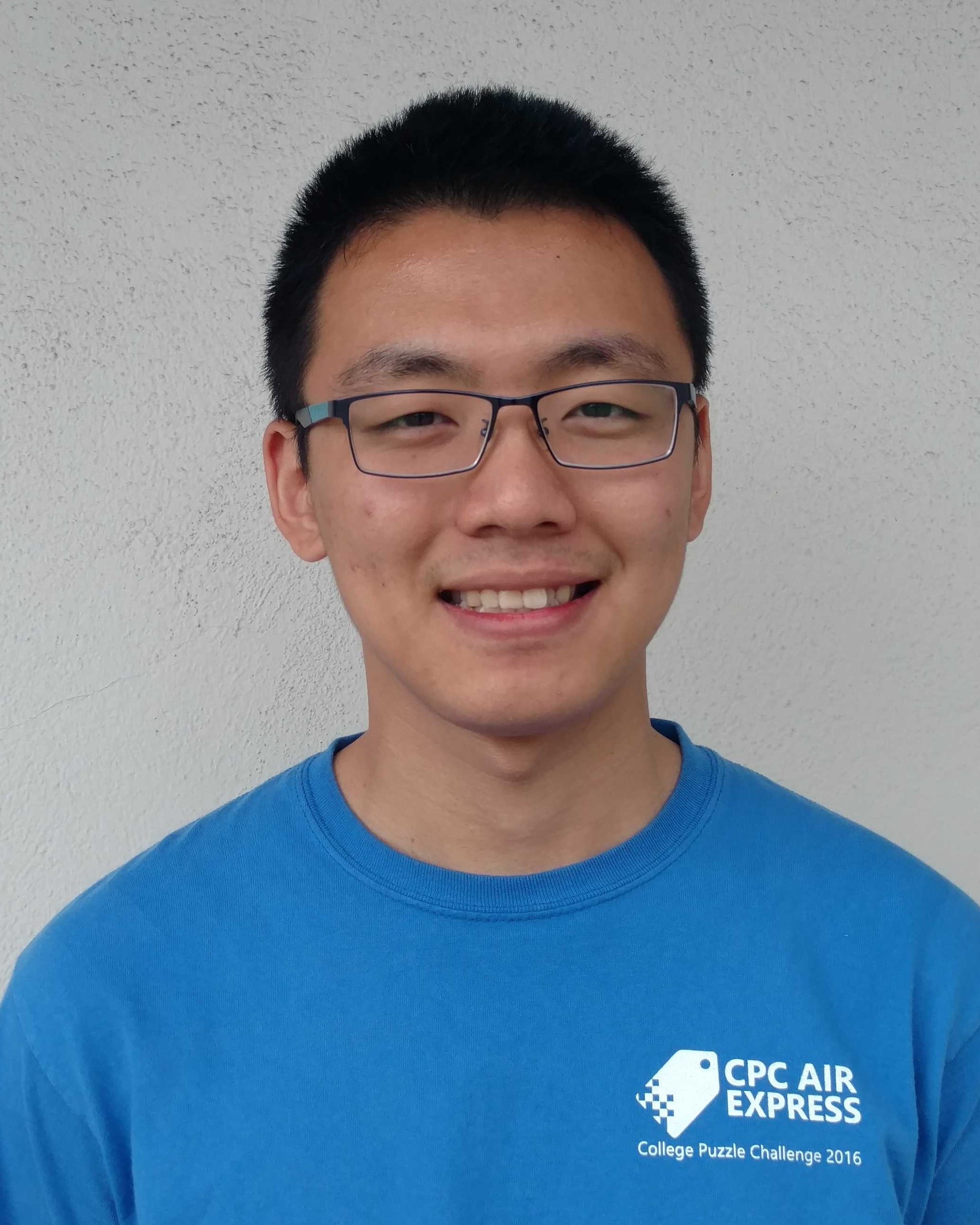}
is a Ph.D. candidate in the Biomimetics and Dexterous Manipulation Lab in the Mechanical Engineering Department at Stanford. He received his B.S. in Mechanical Engineering from Georgia Institute of Technology in 2017. Tony is a NASA Robotics Academy graduate and interned at NASA JPL in the Extreme Environment Robotics group working on the Asteroid Redirect Mission. Tony's research interests include bio-inspired and field robotics, particularly designing climbing and perching robots that interact with challenging, real-world environments. 
\end{biographywithpic}

\begin{biographywithpic}
{Benjamin Hockman}{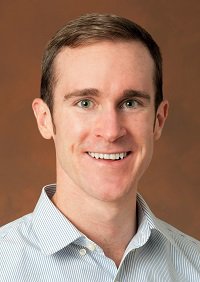}
is a Robotics Technologist at JPL in the Robotic Mobility Group. He received is PhD from Stanford University in 2018, where his graduate work focused on robotic surface mobility on small Solar System bodies. Ben's research interests include design, control, modeling, estimation, and decision making for space robotic systems. Ben has worked on extreme-terrain tethered rovers for exploring Lunar pits, internally-actuated hopping robots for asteroids and comets, melt probes for accessing the oceans of icy moons, and algorithms for spacecraft and rover autonomy.
\end{biographywithpic}

\begin{biographywithpic}
{Edward Schmerling}{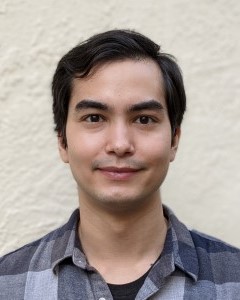}
is a research scientist with the Autonomous Systems Laboratory at Stanford University, where he received his PhD in 2019 in Computational \& Mathematical Engineering. His research interests span the modern robotics stack, with a particular focus on the development of data-driven methodologies for system modeling that may be leveraged to provide quantifiable safety and performance in the face of uncertainty. Prior to returning to Stanford, Ed worked as a research engineer with Waymo Research towards enabling capable and trustworthy autonomous vehicles.
\end{biographywithpic}

\begin{biographywithpic}
{Mark Cutkosky}{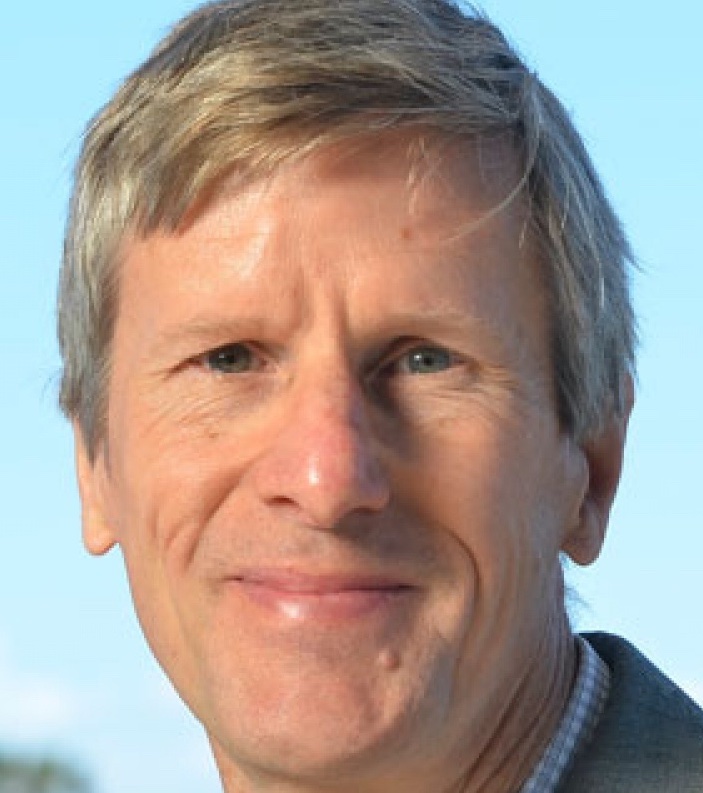}
is the director of the Biomimetics and Dexterous Manipulation Lab in the Mechanical Engineering Department at Stanford. He has been active in bio-inspired robots, dexterous manipulation and rapid prototyping since 1985. His research on insect- and gecko-inspired adhesives resulted in the Spinybot and Stickybot climbing robots, and has been applied to human climbing, perching micro air vehicles and micro robots. Cutkosky is a fellow of IEEE and ASME, a former National Science Foundation Presidential Young Investigator and a former Fulbright Distinguished Chair.
\end{biographywithpic}

\begin{biographywithpic}
{Marco Pavone}{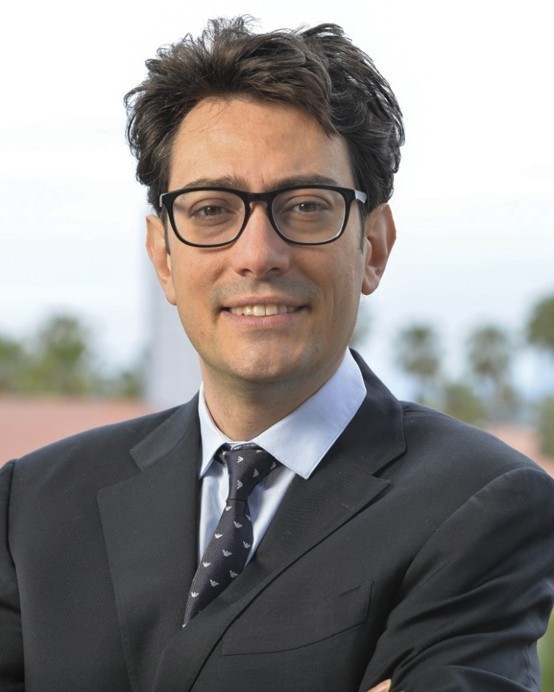}
is an Associate Professor of Aeronautics and Astronautics at Stanford
University, where he is the Director of the Autonomous Systems Laboratory. Before joining Stanford, he was a Research Technologist within the Robotics Section at the NASA Jet Propulsion Laboratory. He received a Ph.D. degree in Aeronautics and Astronautics from the
Massachusetts Institute of Technology in 2010. Dr. Pavone’s expertise lies in the fields of controls and robotics. His main research interests are in the development of methodologies for the analysis, design, and control of autonomous systems, with an emphasis on autonomous aerospace vehicles and large-scale robotic networks.
\end{biographywithpic}
\end{document}